\documentclass{article}
\usepackage{ijcai26}

\usepackage{times}
\usepackage{soul}
\usepackage{url}
\usepackage[hidelinks]{hyperref}
\usepackage[utf8]{inputenc}
\usepackage[small]{caption}
\usepackage{graphicx}
\usepackage{amsmath}
\usepackage{amsthm}
\usepackage{booktabs}
\usepackage{algorithm}
\usepackage{algorithmic}
\usepackage[switch]{lineno}

\usepackage{amsmath,amssymb}      
\usepackage{graphicx}             
\usepackage{subcaption}           
\usepackage{booktabs}             
\usepackage{multirow}             
\usepackage{float}                
\usepackage{array}                
\usepackage{makecell} 

\title{Reducing Bias and Variance: Generative Semantic Guidance and Bi-Layer Ensemble for Image Clustering}

\author{
	Feijiang Li$^{1,2,3}$
	\and
	Zhenxiong Li$^1$
	\and
	Jieting Wang$^{1,2,3}$\\
	Zizheng Jiu$^1$
	\and
	Saixiong Liu$^1$
	\and
	Liang Du$^{1,2,3}$\\
	\affiliations
	$^1$Institute of Big Data Science and Industry, Shanxi University\\
	$^2$Key Laboratory of Evolutionary Science Intelligence of Shanxi Province\\
	$^3$School of Artificial Intelligence, Shanxi University\\
	\emails
	feijiangli@163.com,
	\{lizhenxiong, jtwang, jiuzizheng, duliang\}@sxu.edu.cn,
	liu\_saixiong@126.com
}

\begin{document}

\maketitle

\begin{abstract}

Image clustering aims to partition unlabeled image datasets into distinct groups. A core aspect of this task is constructing and leveraging prior knowledge to guide the clustering process.  Recent approaches introduce semantic descriptions as prior information, most of which typically relying on matching-based techniques with predefined vocabularies. However, the limited matching space restricts their adaptability to downstream clustering tasks. Moreover, these methods primarily focus on reducing bias to improve performance, frequently overlooking the importance of variance reduction. To address these limitations, we propose GSEC (Image Clustering based on Generative Semantic Guidance and Bi-Layer Ensemble), a framework designed to reduce bias through generative semantic guidance and mitigate variance via ensemble learning. Our method employs Multimodal Large Language Models to generate semantic descriptions and derive image embeddings via weighted averaging. Additionally, a bi-layer ensemble strategy integrates cross-modal information through BatchEnsemble in the inner layer and aligns outputs via an alignment mechanism in the outer layer. Comparative experiments demonstrate that GSEC outperforms 18 state-of-the-art methods across six benchmark datasets, while further analysis confirms its effectiveness in simultaneously reducing both bias and variance. The code is available at https://github.com/2017LI/GSEC.git.

\end{abstract}

\section{Introduction}

The goal of image clustering is to partition a dataset into multiple disjoint subsets by learning from unlabeled image data. A core aspect of this task is constructing and leveraging prior knowledge to form effective supervisory signals. 

Based on the nature of the prior information, image clustering methods can be broadly categorized into several types. Classical clustering approaches rely heavily on assumptions about data distributions, including prototype-based methods~\cite{mcqueen1967some}, density-based methods~\cite{ester1996density}, and hierarchical methods~\cite{dasgupta2002performance}. Deep clustering methods leverage powerful representation learning to extract highly discriminative features as prior information~\cite{xie2016unsupervised,caron2018deep}. Advances in self-supervised learning have further shifted clustering paradigms toward supervision signals derived from data augmentation~\cite{li2021contrastive}. Nevertheless, these approaches rely exclusively on intrinsic information within the data to guide clustering. As a result, the supervisory signals are inherently constrained by the data itself~\cite{liu2024interactive}, making further performance improvements increasingly difficult in practice.


Inspired by the zero-shot classification paradigm introduced by CLIP~\cite{radford2021learning}, recent studies have shifted the source of supervision from internal priors to external semantic descriptions. Methods such as SIC~\cite{cai2023semantic} and TAC~\cite{li2023image} typically adopt predefined vocabularies (e.g., WordNet) and employ pre-trained vision–language models to embed images and text into a shared representation space. Then, images are matched with the most relevant textual labels to generate semantic descriptions, which serve as external semantic priors to guide the clustering process. Most of these methods construct semantic prior information by matching images with textual libraries. However, due to the limited coverage of predefined vocabularies, forced alignment between images and text may lead to semantic inconsistency, particularly when complex visual semantics surpass the representational capacity of the fixed lexical set. Therefore, these approaches may be constrained by their reliance on matching-based semantic prior construction.


Moreover, the strategy of introducing semantic priors is essentially to improve the clustering performance by reducing the bias. In machine learning theory, as is well known, estimation error arises from the joint effects of bias, variance, and noise. Therefore, explicitly accounting for variance reduction during clustering is expected to further improve performance.


Based on the discussions above, we propose Image Clustering based on Generative Semantic Guidance and Bi‑Layer Ensemble, noted as GSEC. This framework jointly improves clustering performance by addressing both bias and variance. To reduce bias, we introduce a generative semantic prior construction strategy that overcomes the limitations of matching‑based semantic priors. This strategy leverages Multimodal Large Language Models to generate semantic descriptions and synthesizes corresponding semantic representations for each image via weighted averaging. To reduce variance, we incorporate ensemble learning through a bi‑layer ensemble strategy, comprising an inner‑layer ensemble and an outer‑layer ensemble. The inner layer extracts domain information from both semantic and image modalities and performs internal ensemble via BatchEnsemble to achieve semantically guided image clustering. The outer layer employs an alignment‑based ensemble mechanism to align the inner outputs with the outer outputs, with the aligned result serving as the final clustering outcome.

Our contributions are summarized as follows:

\begin{itemize}
	\item Generative Semantic Guidance: We propose a bias reduction strategy that leverages multimodal large language model to generate adaptive semantic descriptions for images, thereby reducing the inconsistency of the match-based method.
	
	\item Bi‑Layer Ensemble Mechanism: We design a bi-layer ensemble framework to mitigate variance. The inner ensemble integrates multimodal information via BatchEnsemble, and the outer ensemble aligns inner predictions with external outputs, thereby enhancing robustness and stability.
	
	\item Comprehensive Empirical Validation: We conduct experiments on benchmark datasets, demonstrating that GSEC significantly outperforms the reference methods. The bias‑variance decomposition analysis confirms that our approach effectively reduces both bias and variance.
\end{itemize}

\section{Related Works}

In this section, we briefly review image clustering and ensemble learning.

\subsection{Image Clustering}

The evolution of image clustering methodologies has progressively incorporated richer forms of prior knowledge to construct more effective supervisory signals~\cite{ren2024deep}. In the following, we review these methods according to their primary sources of supervision: internal prior information and external semantic information.

The development of the internal prior information-based clustering algorithms can be broadly divided into three stages. Early classical clustering methods typically rely on low-level visual semantics: K-means~\cite{mcqueen1967some} groups data by minimizing distances to cluster centers, while DBSCAN~\cite{ester1996density} identifies arbitrarily shaped clusters based on local density. With the advent of deep learning, deep clustering methods emerged: DEC~\cite{xie2016unsupervised} learns feature representations via autoencoders and jointly optimizes a clustering objective; ClusterGAN~\cite{mukherjee2019clustergan} implements clustering through adversarial learning. Subsequently, self-supervised approaches were developed: DeepCluster~\cite{caron2018deep} iteratively uses clustering outcomes as pseudo-labels for training, and CC~\cite{li2021contrastive} further integrates contrastive learning with clustering objectives to enhance performance.


Vision-language pre-training models such as CLIP~\cite{radford2021learning} have shifted image clustering from relying on internal priors to exploiting external semantic information. Based on CLIP, ZOC~\cite{esmaeilpour2022zero} performs zero-shot recognition of unknown classes, while SIC~\cite{cai2023semantic} and TAC~\cite{li2023image} enhance clustering by modeling image-text geometric consistency and incorporating WordNet semantics with cross-modal distillation, respectively. GradNorm~\cite{peng2025provable} further improves text-assisted image clustering by selecting semantically relevant nouns from unlabeled data.~\cite{wang2026text} extend text-guided learning to visual adaptation through deep manifold constraints and neighborhood propagation.

Despite these advancements, existing multimodal clustering methods remain fundamentally limited by their reliance on matching mechanisms within a constrained vocabulary space. Due to the restricted expressiveness of predefined vocabularies, forced alignment between images and textual labels inevitably leads to semantic inconsistency when complex visual semantics exceed the coverage of the fixed lexical set.

Moreover, existing strategies that incorporate semantic priors primarily target bias reduction to improve clustering. It is important to note that estimation error stems from the combined effects of bias, variance, and noise. Therefore, to achieve more robust and stable clustering, addressing variance reduction is also essential. Ensemble learning serves as a key technique for variance reduction in machine learning.

\subsection{Ensemble Learning}

Ensemble learning accomplishes complex tasks by constructing and combining multiple learners~\cite{dietterich2000ensemble}. Breiman~\cite{breiman1996bagging} demonstrated that generating various versions of a predictor and aggregating their outputs can substantially reduce model variance.

Traditional ensemble methods are broadly categorized into two paradigms: Bagging and Boosting. Bagging~\cite{breiman1996bagging} promotes diversity by creating multiple training subsets via bootstrap sampling and independently training the same learning algorithm on each subset; Random Forest~\cite{breiman2001random} is a representative example. In contrast, Boosting employs a sequential training strategy, where each subsequent learner is explicitly optimized to correct errors made by the preceding ensemble, with AdaBoost~\cite{freund1997decision} being a canonical instance. 

\begin{figure*}[!ht]
	\centering
    \includegraphics[width=1.0\linewidth, height=0.4\textheight]{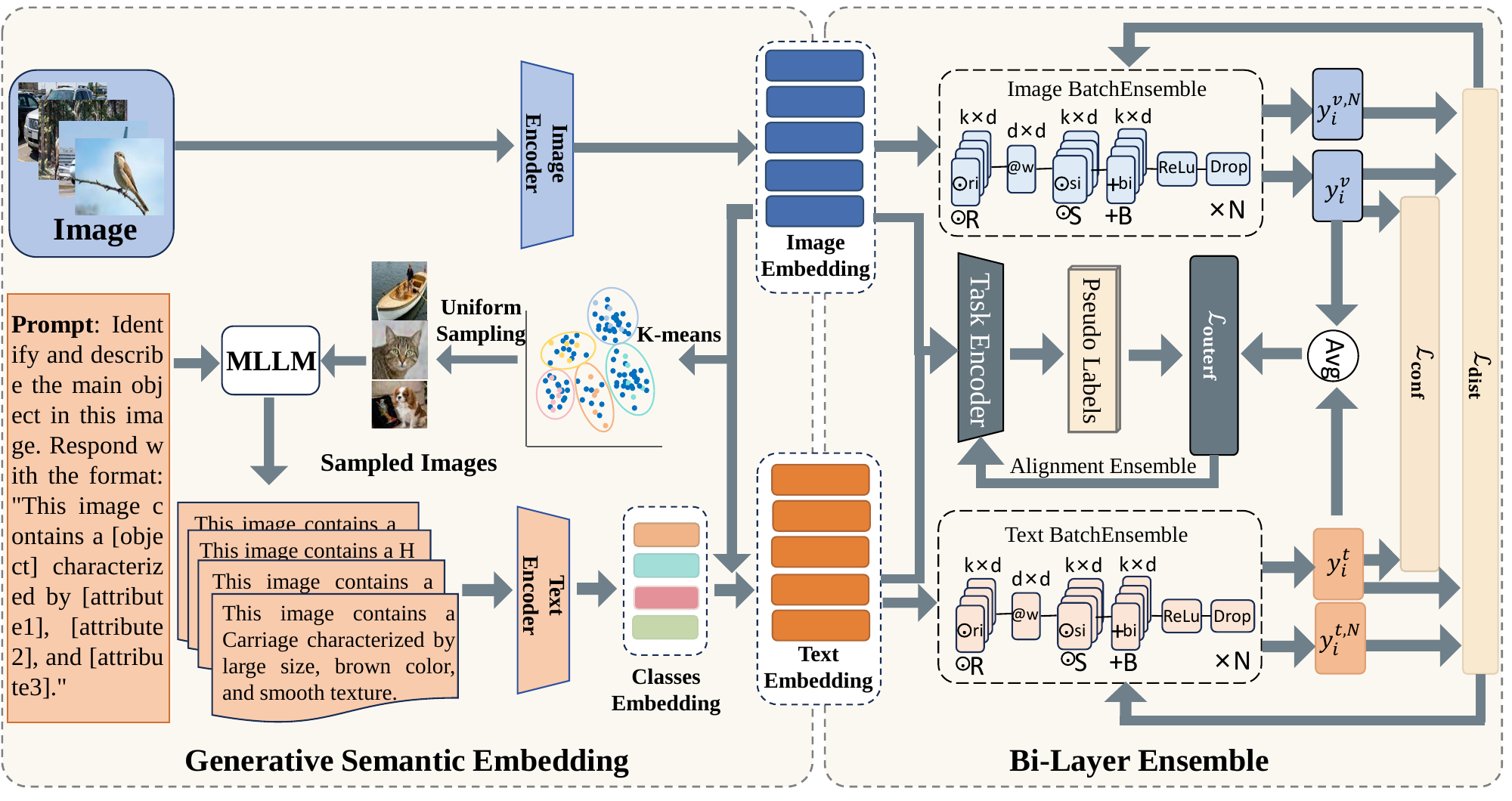}
	\caption{
	\textbf{Overall Framework of GSEC.} The framework integrates generative semantic embedding with a bi‑layer ensemble strategy.
	}
	\label{fig:framework}
\end{figure*}


Building upon classical ensemble principles, deep learning methods have incorporated Bagging- and Boosting-inspired strategies to improve robustness and generalization. For example, BatchEnsemble improves efficiency through rank-one parameter sharing with minimal additional parameters~\cite{wen2020batchensemble}. Recent studies further explore selective ensemble learning based on pure accuracy~\cite{wang2022generalization} and clustering ensemble methods that aggregate multiple base partitions into robust consensus results, including GOT~\cite{li2021got}, fuzzy ensemble clustering~\cite{li2023fuzzy}, and k-HyperEdge medoids~\cite{li2025k}.

In this work, we introduce ensemble methods to reduce variance in image clustering.

\section{Method}

In this section, we propose an image clustering method based on generative semantic guidance and bi-Layer ensemble (GSEC), with its overall framework illustrated in Figure~\ref{fig:framework}. Firstly, we introduce the generation of semantic embeddings using Multimodal Large Language Models. Secondly, we elaborate on the operational mechanism of the designed bi-layer ensemble framework.

\subsection{Generative Semantic Embedding}

To address the limitation of constrained semantic matching space, we leverage Multimodal Large Language Models (MLLMs) to generate semantically enriched descriptions through reasoning. The procedure is detailed as follows.

Let $\mathcal{V}=\{I_i\}_{i=1}^n$ denote the image set, where $n$ is the number of samples. Using an image encoder \(E_v\), we first extract image embeddings \(v_i = E_v(I_i) \in \mathbb{R}^d\), with \(d\) representing the embedding dimensionality.

To uncover latent semantic structures, we perform K-means clustering on all image embeddings \(\{v_i\}_{i=1}^n\). Following the practice in \cite{peng2025provable,li2023image}, the number of clusters is set as $C= \max\left\{ \frac{n}{300},\; 3K \right\}$, where \(K\) denotes the expected number of clusters. Images within each cluster are then sorted by the distance between their embeddings and the corresponding cluster center. From each cluster, we uniformly select up to five representative images according to this sorted order. If a cluster contains fewer than five images, all of them are selected.

These selected images are fed into the Multimodal Large Language Model using a natural-language prompt designed to capture core visual semantics. The prompt is:" Identify and describe the main object in this image. Respond with the format:' This image contains a [object] characterized by [attribute1], [attribute2], and [attribute3]'".

The generated class-level descriptions \(T_k\) are encoded by the CLIP text encoder \(E_t\) into class embeddings \(\bar{t}_j = E_t(T_k)\). The similarity between each image embedding \(v_i\) and every class text embedding \(\{\bar{t}_j\}_{j=1}^M\) is then computed to derive an image-specific text embedding \(t_i\), where \(M\) is the number of selected samples.

Specifically, we first compute the conditional probability of the \(j\)-th class text embedding given image \(v_i\):
\begin{equation}
	p(\bar{t}_j|v_i) = \frac{\exp(\text{sim}(v_i, \bar{t}_j)/\tilde{\tau})}{\sum_{k=1}^M \exp(\text{sim}(v_i, \bar{t}_k)/\tilde{\tau})},
\end{equation}
where $\text{sim}(\cdot, \cdot)$ denotes cosine similarity, and $\tilde{\tau}$ is a temperature parameter that controls the smoothness of the distribution. Subsequently, the final text counterpart $t_i$ is obtained as a weighted average of all classes embedding:
\begin{equation}
	t_i = \sum_{j=1}^M p(\bar{t}_j|v_i)\bar{t}_j.
\end{equation}
Here, $t_i$ represents the text embedding of the $i$-th sample.

\subsection{Bi-Layer Ensemble}

To mitigate variance, we design a bi-layer ensemble framework: the inner layer employs a BatchEnsemble integrator for cross-modal clustering, and the outer layer further aligns and integrates the inner outputs with external predictions through an alignment-based ensemble mechanism.

\subsubsection{Inner Ensemble}

At the inner layer, we construct an integrator based on the BatchEnsemble method to efficiently integrate cross-modal data. This integrator consists of two independent branches for image and text modalities, each comprising a set of shared weights and member-specific low-rank modulation parameters.

The output of the $k$-th base model in the BatchEnsemble integrator is given by:
\begin{equation}
	l_k(x) = s_k \odot (W(r_k \odot x)) + b_k,
\end{equation}
where $W$ denotes the shared weight matrix across all members, $r_k$ and $s_k$ are learnable rank-1 vectors specific to the $k$-th member, $\odot$ represents the element-wise multiplication, and $b_k$ is the bias term for the $k$-th member.

We denote the BatchEnsemble integrators for the image and text modalities as \(l_k^v(x)\) and \(l_k^t(x)\), respectively. Let \(v_i^N\) be a random neighbor of \(v_i\), and \(t_i^N\) be a random neighbor of \(t_i\). We process \(v_i\) and \(v_i^N\) through the image BatchEnsemble integrator, and \(t_i\) and \(t_i^N\) through the text BatchEnsemble integrator. As for $y_i^v$ and $y_i^t$:
\begin{equation}
	y_i^v = \frac{1}{m} \sum_{k=1}^{m} l_k^v(v_i),\quad  y_i^t = \frac{1}{m} \sum_{k=1}^{m} l_k^t(t_i),
\end{equation}
where $m$ is the ensemble size, $y_i^v$ and $y_i^t$ represent the soft assignments for the $i$-th image and text, respectively. As for $y_i^{v,N}$ and $y_i^{t,N}$: 
\begin{equation}
	y_i^{v,N} = \frac{1}{m} \sum_{k=1}^{m} l_k^v(v_i^N),\quad y_i^{t,N} = \frac{1}{m} \sum_{k=1}^{m} l_k^t(t_i^N),
\end{equation}
where $y_i^{v,N}$ and $y_i^{t,N}$ denote the neighbor soft assignments for the $i$-th image and text, respectively.

To achieve mutual guidance between the text and image modalities during clustering, we introduce the Cross-Modal Mutual Distillation loss:
\begin{equation}
	\mathcal{L}_{\mathrm{dist}} = \sum_{i=1}^n \left( D_{\mathrm{KL}}(y_i^t \| y_i^{v,N}) + D_{\mathrm{KL}}(y_i^v \| y_i^{t,N}) \right),
\end{equation}
where \( D_{\mathrm{KL}} \) denotes the KL divergence. This loss serves a dual purpose: it enhances the discriminability of cluster boundaries by reducing inter-cluster similarity, while also enabling bidirectional alignment between each image and the neighbors of its corresponding text, and between each text and the neighbors of its corresponding image.

To stabilize training and encourage the model to produce more confident cluster assignments, we introduce the confidence loss:
\begin{equation}
	\mathcal{L}_{\mathrm{conf}} = -\log \sum_{i=1}^n (y_i^v)^\top y_i^t.
\end{equation}

%

To prevent the collapse of all samples into only a few clusters, we adopt the balance loss:
\begin{equation}
	\mathcal{L}_{\mathrm{bal}} = -\sum_{i=1}^K \left( \bar{y^v} \log \bar{y^v} + \bar{y^t} \log \bar{y^t} \right),
\end{equation}
where \(K\) is the number of clusters. \(\bar{y}^v\) and \(\bar{y}^t\) denote the cluster assignment distributions in the image and text modalities, respectively, which are computed as:
\begin{equation}
\bar{y^t} = \frac{1}{n} \sum_{i=1}^n y_i^t,\quad\bar{y^v} = \frac{1}{n} \sum_{i=1}^n y_i^v.
\end{equation}

%

Then, the inner ensemble aims to optimize the model parameters by minimizing the following comprehensive loss function:
\begin{equation}
	\mathcal{L}_{\text{inner}} = \mathcal{L}_{\text{dist}} + \mathcal{L}_{\text{conf}} - \mathcal{L}_{\text{bal}}.
\end{equation}

\subsubsection{Outer Ensemble}

After the inner-layer ensemble training converges, we compute the average of the outputs from the image and text modalities:
\begin{equation}
	\hat{y}_i = \frac{1}{2}(y_i^v + y_i^t),
\end{equation}
where \(\hat{y}_i\) denotes the output of the inner ensemble.

In the outer ensemble stage, we train a task encoder \(\phi\) that takes the concatenated image and text embeddings \([v_i; t_i]\) as input, with its output defined as:
\begin{equation}
	y_i = \phi\left([v_i; t_i]\right).
\end{equation}

To align the predictions of the inner ensemble with the outputs of the outer task encoder, we adopt the cross-entropy loss as the alignment objective:
\begin{equation}
	\mathcal{L}_{\text{align}} = \sum_{i=1}^{n} \mathcal{L}_{\text{CE}}(y_i, \hat{y}_i).
\end{equation}

The outer ensemble optimizes the parameters \(\phi\) by minimizing the following comprehensive loss:
\begin{equation}
	\mathcal{L}_{\text{outer}} = \mathcal{L}_{\text{align}} -  H(\bar{p}),
\end{equation}
where $\bar{p} = \frac{1}{n}\sum_{i=1}^{n} \phi([\mathbf{v}_i; \mathbf{t}_i])$ is the average prediction distribution, and \(H(\bar{p})\) denotes its entropy.


After the outer ensemble training converges, the final cluster assignment for each sample is obtained from the output of the task encoder. This decision mechanism integrates the multimodal alignment information from the inner ensemble outputs with the comprehensive inference of the task encoder, which is expected to generate a stable and consistent clustering result.


 
\section{Experiments}

In this section, we validate GSEC's performance through benchmark comparisons. Additionally, we conduct bias-variance decomposition analysis to verify the effectiveness of our method, along with ablation studies, hyperparameter analysis, convergence analysis, and visualization analysis (In Appendix~\ref{app:Visualized results}).

\subsection{Experiment Setup}

\paragraph{Datasets} To evaluate the effectiveness of our proposed method, we conduct experiments on 11 widely used benchmark datasets. The detailed information of these datasets is summarized in Table \ref{tab:benchmark_datasets}.

\begin{table}[htbp]
	\centering
	\caption{\textbf{Statistics of the benchmark datasets.} This table summarizes the details of the eleven datasets used in the experiments.}
	\renewcommand{\arraystretch}{1.2}
    \small
	\setlength{\tabcolsep}{3pt} 
	\begin{tabular}{@{}ccccc@{}}
		\toprule
		No & Dataset & Cls & \makecell[c]{Train\\size} & \makecell[c]{Test\\size} \\
		\midrule
		1  & CIFAR10~\cite{krizhevsky2009learning} & 10 & 50,000 & 10,000 \\
		2  & CIFAR100~\cite{krizhevsky2009learning} & 100 & 50,000 & 10,000 \\
		3  & STL10~\cite{coates2011analysis} & 10 & 5,000 & 8,000 \\
		4  & ImageNet-10~\cite{chang2017deep} & 10 & 13,000 & 500 \\
		5  & ImageNet-Dogs~\cite{chang2017deep} & 15 & 19,500 & 750 \\
		6  & Food101~\cite{bossard2014food} & 101 & 75,750 & 25,250 \\
		7  & StanfordCars~\cite{krause20133d} & 196 & 8,144 & 8,041 \\
		8  & OxfordPets~\cite{parkhi2012cats} & 37 & 3,680 & 3,669 \\
		9  & FGVC Aircraft~\cite{maji2013fine} & 100 & 6,667 & 3,333 \\
		10 & Country211~\cite{radford2021learning} & 211 & 42,200 & 21,100 \\
		11 & ImageNet-1K~\cite{deng2009imagenet} & 1000 & 1,281,167 & 50,000 \\
		\bottomrule
	\end{tabular}
	\label{tab:benchmark_datasets}
\end{table}
\paragraph{Evaluation}
To evaluate the clustering performance, we employ three widely adopted metrics: Accuracy (ACC), Normalized Mutual Information (NMI), and Adjusted Rand Index (ARI). For all three metrics, higher values indicate better performance.

\paragraph{Implementation Details}
We utilize the pre-trained CLIP-ViT-B/32 model to extract image and text embeddings. For semantic descriptions, we employ Llama-3.2-Vision-11B to generate descriptions from representative images, which are then standardized into a unified format. The learning rates for both the internal integrator and the task encoder are set to 0.001, and the number of ensemble is set to 24. The batch size is fixed at 1024, with the exception of ImageNet-1K, which uses a batch size of 8192. All experiments were conducted on a single NVIDIA A6000 GPU.

\subsection{Comparison with Baseline Methods}

To comprehensively assess the effectiveness of GSEC, we benchmark it against a wide range of state-of-the-art algorithms, categorized into two groups: (1) Unimodal Clustering Methods, including DEC~\cite{xie2016unsupervised}, DAC~\cite{chang2017deep}, JULE~\cite{yang2016joint}, DCCM \cite{wu2019deep}, IIC~\cite{ji2019invariant}, SCAN~\cite{van2020scan}, TCL~\cite{tao2021clustering}, TCC~\cite{shen2021you}, GCC~\cite{zhong2021graph}, CRLC~\cite{do2021clustering}, RPSC~\cite{liu2024rpsc}, PICA~\cite{huang2020deep}, SPICE~\cite{niu2022spice}, DivClust ~\cite{metaxas2023divclust}, and LFSS~\cite{li2025learning}; (2) Multimodal Clustering Methods, such as TAC~\cite{li2023image}, CLIP (K-means), SIC~\cite{cai2023semantic}, GradNorm ~\cite{peng2025provable}, and CAE~\cite{zhu2026hierarchical}.


We evaluate our proposed GSEC on the first five datasets listed in Table \ref{tab:benchmark_datasets} and compare it with the above mentioned 18 image clustering baselines. As shown in Table~\ref{tab:sota_comparison}, our method consistently outperforms these competitors. Furthermore, we test GSEC on the challenging large‑scale ImageNet‑1K dataset and compare it with four leading multimodal methods. As presented in Table~\ref{tab:performance_comparison}, our approach retains its superiority and delivers the best performance.

\begin{table*}[!t]
    \centering
    \caption{\textbf{Comparison with reference baseline methods.} The best results are highlighted in \textbf{bold}.}
    \label{tab:sota_comparison}
   \scriptsize
   \setlength{\tabcolsep}{5pt}
   \renewcommand{\arraystretch}{0.6}
    \resizebox{\textwidth}{!}{
        \begin{tabular}{l ccc ccc ccc ccc ccc}
            \toprule
            Dataset & \multicolumn{3}{c}{CIFAR10} & \multicolumn{3}{c}{CIFAR100} & \multicolumn{3}{c}{ImageNet-10} & \multicolumn{3}{c}{ImageNet-Dogs} & \multicolumn{3}{c}{STL10} \\
            \cmidrule(lr){2-4} \cmidrule(lr){5-7} \cmidrule(lr){8-10} \cmidrule(lr){11-13} \cmidrule(lr){14-16}
            Metrics & NMI & ACC & ARI & NMI & ACC & ARI & NMI & ACC & ARI & NMI & ACC & ARI & NMI & ACC & ARI \\
            \midrule
            JULE (CVPR16)      & 19.2 & 27.2 & 13.8 & 10.3 & 13.7 & 3.3  & 17.5 & 30.0 & 13.8 & 5.4  & 13.8 & 2.8  & 18.2 & 27.7 & 16.4 \\
            DEC (ICML16)       & 25.7 & 30.1 & 16.1 & 13.6 & 18.5 & 5.0  & 28.2 & 38.1 & 20.3 & 12.2 & 19.5 & 7.9  & 27.6 & 35.9 & 18.6 \\
            DAC (ICCV17)       & 39.6 & 52.2 & 30.6 & 18.5 & 23.8 & 8.8  & 39.4 & 52.7 & 30.2 & 21.9 & 27.5 & 11.1 & 36.6 & 47.0 & 25.7 \\
            DCCM (ICCV19)      & 49.6 & 62.3 & 40.8 & 28.5 & 32.7 & 17.3 & 60.8 & 71.0 & 55.5 & 32.1 & 38.3 & 18.2 & 37.6 & 48.2 & 26.2 \\
            IIC (ICCV19)       & 51.3 & 61.7 & 41.1 & 22.5 & 25.7 & 11.7 & --   & --   & --   & --   & --   & --   & 49.6 & 59.6 & 39.7 \\
            PICA (CVPR20)      & 59.1 & 69.6 & 51.2 & 31.0 & 33.7 & 17.1 & 80.2 & 87.0 & 76.1 & 35.2 & 35.3 & 20.1 & 61.1 & 71.3 & 53.1 \\
            SCAN (ECCV20)      & 79.7 & 88.3 & 77.2 & 48.6 & 50.7 & 33.3 & --   & --   & --   & 61.2 & 59.3 & 45.7 & 69.8 & 80.9 & 64.6 \\
            GCC (ICCV21)       & 76.4 & 85.6 & 72.8 & 47.2 & 47.2 & 30.5 & 84.2 & 90.1 & 82.2 & 49.0 & 52.6 & 36.2 & 68.4 & 78.8 & 63.1 \\
            CRLC (ICCV21)      & 67.9 & 79.9 & 63.4 & 41.6 & 42.5 & 26.3 & 83.1 & 85.4 & 75.9 & 48.4 & 46.1 & 59.7 & 72.9 & 81.8 & 62.8 \\
            TCC (NeurIPS21)    & 79.0 & 90.6 & 73.3 & 47.9 & 49.1 & 31.2 & 84.8 & 89.7 & 82.5 & 55.4 & 59.5 & 41.7 & 72.2 & 81.4 & 68.9 \\
            TCL (IJCV22)       & 81.9 & 88.7 & 78.0 & 52.9 & 53.1 & 35.7 & 87.5 & 89.5 & 83.7 & 62.3 & 64.4 & 51.6 & 79.9 & 86.8 & 75.7 \\
            SPICE (TIP22)      & 73.4 & 83.8 & 70.5 & 44.8 & 46.8 & 29.4 & 82.8 & 92.1 & 83.6 & 57.2 & 64.6 & 68.7 & 81.7 & 90.8 & 81.2 \\
            DivClust (CVPR23)  & 71.0 & 81.5 & 67.5 & 44.0 & 43.7 & 28.3 & 85.0 & 90.0 & 81.9 & 51.6 & 52.9 & 37.6 & --   & --   & --   \\
            RPSC (AAAI24)      & 75.4 & 85.7 & 73.1 & 47.6 & 51.8 & 34.1 & 83.0 & 92.7 & 85.8 & 55.2 & 64.0 & 46.5 & 83.8 & 92.0 & 83.4 \\
            LFSS (ICML25)      & 87.2 & 93.4 & 86.8 & --   & --   & --   & 85.6 & 93.2 & 85.7 & 61.6 & 69.1 & 53.3 & 77.1 & 86.1 & 74.0 \\
            \midrule
            CLIP (k-means)     & 70.3 & 74.2 & 61.6 & 49.9 & 45.5 & 28.3 & 96.9 & 98.2 & 96.1 & 39.8 & 38.1 & 20.1 & 91.7 & 94.3 & 89.1 \\
            SIC (AAAI23)       & 84.7 & 92.6 & 84.4 & 59.3 & 58.3 & 43.9 & 97.0 & 98.2 & 96.1 & 69.0 & 69.7 & 55.8 & 95.3 & 98.1 & 95.9 \\
            TAC (ICML24)       & 83.3 & 91.9 & 83.1 & 61.1 & 60.7 & 44.8 & 98.5 & 99.2 & 98.3 & 80.6 & 83.0 & 72.2 & 95.5 & 98.2 & 96.1 \\
            GradNorm (IEEE25)  & 82.6 & 91.1 & 81.5 & 67.2 & 54.8 & 37.0 & 98.7 & 99.4 & 98.7 & 81.0 & 81.2 & 70.9 & 95.7 & 98.3 & 96.2 \\
            CAE (AAAI26)       & 81.7 & 90.9 & 80.5 & --   & --   & --   & 98.1 & 98.8 & 97.4 & 77.9 & 77.5 & 66.3 & 95.5 & 98.2 & 96.7 \\
            \midrule
            \textbf{GSEC}      & \textbf{91.2} & \textbf{96.3} & \textbf{92.1} & \textbf{71.8} & \textbf{61.8} & \textbf{48.0} & \textbf{99.6} & \textbf{99.8} & \textbf{99.6} & \textbf{86.3} & \textbf{87.1} & \textbf{79.2} & \textbf{97.6} & \textbf{99.4} & \textbf{97.7} \\
            \bottomrule
        \end{tabular}
    }
\end{table*}

\begin{table*}[!t]
    \centering
    \caption{\textbf{Effectiveness Analysis of Semantic Prior Information.} The table reports the clustering accuracy achieved using semantic priors obtained from WordNet matching (M‑Text) and generative descriptions (G‑Text).}
    \label{tab:multimodal_ablation}
    \small
    \resizebox{\textwidth}{!}{
        \begin{tabular}{l|ccc|ccc|ccc|ccc|ccc}
            \toprule
            \multirow{2}{*}{\textbf{Backbone}} & \multicolumn{3}{c|}{\textbf{CIFAR10}} & \multicolumn{3}{c|}{\textbf{CIFAR100}} & \multicolumn{3}{c|}{\textbf{ImageNet-Dogs}} & \multicolumn{3}{c|}{\textbf{STL10}} & \multicolumn{3}{c}{\textbf{ImageNet-10}} \\
            \cmidrule(lr){2-4} \cmidrule(lr){5-7} \cmidrule(lr){8-10} \cmidrule(lr){11-13} \cmidrule(lr){14-16}
            & M-Text & G-Text & \textbf{Gain} & M-Text & G-Text & \textbf{Gain} & M-Text & G-Text & \textbf{Gain} & M-Text & G-Text & \textbf{Gain} & M-Text & G-Text & \textbf{Gain} \\
            \midrule
            CLIP-RN50     & 71.56 & 72.10 & \textbf{0.54} & 31.59 & 34.87 & \textbf{3.28} & 77.82 & 78.80 & \textbf{0.98} & 61.46 & 96.70 & \textbf{35.24} & 98.40 & 98.52 & \textbf{0.12} \\
            CLIP-RN101    & 81.74 & 81.83 & \textbf{0.09} & 40.47 & 44.50 & \textbf{4.03} & 76.93 & 82.93 & \textbf{6.00} & 76.71 & 98.11 & \textbf{21.40} & 98.60 & 98.63 & \textbf{0.03} \\
            CLIP-RN50x4   & 76.39 & 78.24 & \textbf{1.85} & 38.41 & 38.65 & \textbf{0.24} & 82.40 & 83.33 & \textbf{0.93} & 70.35 & 97.92 & \textbf{27.57} & 98.20 & 98.80 & \textbf{0.60} \\
            CLIP-ViT-B/32 & 89.58 & 90.16 & \textbf{0.58} & 46.41 & 49.34 & \textbf{2.93} & 77.20 & 78.67 & \textbf{1.47} & 73.90 & 98.32 & \textbf{24.42} & 98.20 & 98.72 & \textbf{0.52} \\
            \bottomrule
            \toprule
            \multirow{2}{*}{\textbf{Backbone}} & \multicolumn{3}{c|}{\textbf{StanfordCars}} & \multicolumn{3}{c|}{\textbf{OxfordPets}} & \multicolumn{3}{c|}{\textbf{Aircraft}} & \multicolumn{3}{c|}{\textbf{Country211}} & \multicolumn{3}{c}{\textbf{Food101}} \\
            \cmidrule(lr){2-4} \cmidrule(lr){5-7} \cmidrule(lr){8-10} \cmidrule(lr){11-13} \cmidrule(lr){14-16}
            & M-Text & G-Text & \textbf{Gain} & M-Text & G-Text & \textbf{Gain} & M-Text & G-Text & \textbf{Gain} & M-Text & G-Text & \textbf{Gain} & M-Text & G-Text & \textbf{Gain} \\
            \midrule
            CLIP-RN50     & 30.52 & 50.22 & \textbf{19.70} & 59.42 & 79.50 & \textbf{20.08} & 19.20 & 24.42 & \textbf{5.22} & 9.12 & 12.37 & \textbf{3.25} & 54.54 & 74.96 & \textbf{20.42} \\
            CLIP-RN101    & 28.04 & 56.22 & \textbf{28.18} & 62.20 & 82.42 & \textbf{20.22} & 22.26 & 27.62 & \textbf{5.36} & 9.65 & 13.67 & \textbf{4.02} & 63.18 & 78.38 & \textbf{15.20} \\
            CLIP-RN50x4   & 34.57 & 61.45 & \textbf{26.88} & 70.37 & 87.49 & \textbf{17.12} & 25.92 & 30.40 & \textbf{4.48} & 10.41 & 15.36 & \textbf{4.95} & 67.30 & 78.77 & \textbf{11.47} \\
            CLIP-ViT-B/32 & 37.33 & 50.69 & \textbf{13.36} & 61.52 & 73.83 & \textbf{12.31} & 15.99 & 26.53 & \textbf{10.54} & 9.22 & 13.06 & \textbf{3.84} & 66.90 & 78.22 & \textbf{11.32} \\
            \bottomrule
        \end{tabular}
    }
\end{table*}

\begin{table*}[!t]
    \centering
    \caption{\textbf{Effectiveness of the bi‑layer ensemble strategy across different backbones.} The table compares the clustering accuracy of the single‑image baseline (Image) and the ensemble‑based method (Ensemble) on 10 downstream tasks.}
    \label{tab:ensemble_ablation}
    \small
    \resizebox{\textwidth}{!}{
        \begin{tabular}{l|ccc|ccc|ccc|ccc|ccc}
            \toprule
            \multirow{2}{*}{\textbf{Backbone}} & \multicolumn{3}{c|}{\textbf{CIFAR10}} & \multicolumn{3}{c|}{\textbf{CIFAR100}} & \multicolumn{3}{c|}{\textbf{ImageNet-Dogs}} & \multicolumn{3}{c|}{\textbf{STL10}} & \multicolumn{3}{c}{\textbf{ImageNet-10}} \\
            \cmidrule(lr){2-4} \cmidrule(lr){5-7} \cmidrule(lr){8-10} \cmidrule(lr){11-13} \cmidrule(lr){14-16}
            & Image & Ensemble & \textbf{Gain} & Image & Ensemble & \textbf{Gain} & Image & Ensemble & \textbf{Gain} & Image & Ensemble & \textbf{Gain} & Image & Ensemble & \textbf{Gain} \\
            \midrule
            CLIP-RN50 & 57.20 & 62.44 & \textbf{5.24} & 30.80 & 31.23 & \textbf{0.43} & 45.20 & 47.07 & \textbf{1.87} & 96.60 & 96.69 & \textbf{0.09} & 98.20 & 98.40 & \textbf{0.20} \\
            CLIP-RN101 & 71.60 & 74.68 & \textbf{3.08} & 40.20 & 41.42 & \textbf{1.22} & 51.13 & 66.27 & \textbf{15.14} & 98.00 & 98.02 & \textbf{0.02} & 98.20 & 98.60 & \textbf{0.40} \\
            CLIP-RN50x4 & 67.70 & 69.90 & \textbf{2.20} & 37.70 & 37.83 & \textbf{0.13} & 59.47 & 74.13 & \textbf{14.66} & 97.80 & 97.89 & \textbf{0.09} & 98.60 & 98.80 & \textbf{0.20} \\
            CLIP-ViT-B/32 & 85.90 & 86.19 & \textbf{0.29} & 45.80 & 46.61 & \textbf{0.81} & 48.40 & 63.60 & \textbf{15.20} & 98.30 & 98.31 & \textbf{0.01} & 98.40 & 98.60 & \textbf{0.20} \\
            \bottomrule
            \toprule
            \multirow{2}{*}{\textbf{Backbone}} & \multicolumn{3}{c|}{\textbf{StanfordCars}} & \multicolumn{3}{c|}{\textbf{OxfordPets}} & \multicolumn{3}{c|}{\textbf{Aircraft}} & \multicolumn{3}{c|}{\textbf{Country211}} & \multicolumn{3}{c}{\textbf{Food101}} \\
            \cmidrule(lr){2-4} \cmidrule(lr){5-7} \cmidrule(lr){8-10} \cmidrule(lr){11-13} \cmidrule(lr){14-16}
            & Image & Ensemble & \textbf{Gain} & Image & Ensemble & \textbf{Gain} & Image & Ensemble & \textbf{Gain} & Image & Ensemble & \textbf{Gain} & Image & Ensemble & \textbf{Gain} \\
            \midrule
            CLIP-RN50 & 37.00 & 38.85 & \textbf{1.85} & 51.20 & 54.10 & \textbf{2.90} & 23.00 & 24.06 & \textbf{1.06} & 8.90 & 9.39 & \textbf{0.49} & 65.00 & 67.73 & \textbf{2.73} \\
            CLIP-RN101 & 46.30 & 49.67 & \textbf{3.37} & 66.70 & 71.49 & \textbf{4.79} & 24.20 & 27.36 & \textbf{3.16} & 9.50 & 9.65 & \textbf{0.15} & 74.90 & 75.21 & \textbf{0.31} \\
            CLIP-RN50x4 & 52.90 & 53.23 & \textbf{0.33} & 74.30 & 74.65 & \textbf{0.35} & 26.70 & 30.27 & \textbf{3.57} & 10.10 & 10.41 & \textbf{0.31} & 80.70 & 81.24 & \textbf{0.54} \\
            CLIP-ViT-B/32 & 42.30 & 44.09 & \textbf{1.79} & 62.10 & 63.91 & \textbf{1.81} & 24.00 & 26.01 & \textbf{2.01} & 9.80 & 9.83 & \textbf{0.03} & 72.40 & 73.78 & \textbf{1.38} \\
            \bottomrule
        \end{tabular}
    }
\end{table*}

\begin{table}[t]
	\centering
	\caption{\textbf{Comparison on the large-scale ImageNet-1K dataset.}}
	\label{tab:performance_comparison}
	\small 
	\renewcommand{\arraystretch}{1.0} 
\begin{tabular*}{\hsize}{@{\extracolsep{\fill}}lccccc}
		\toprule
		Metric & CLIP (k-means) & SIC & TAC & GradNorm & \textbf{GSEC} \\
		\midrule
		NMI & 72.3 & 77.2 & 77.8 & 79.2 & \textbf{81.3} \\
		ACC & 38.9 & 47.0 & 48.9 & 52.6 & \textbf{62.5} \\
		ARI & 27.1 & 34.3 & 36.4 & 39.1 & \textbf{52.8} \\
		\bottomrule
	\end{tabular*}
\end{table}

\subsection{Reduction of Bias-Variance Analysis}

\begin{figure*}[t]
	\centering
	\includegraphics[width=1.0\linewidth,height=0.2\textheight]{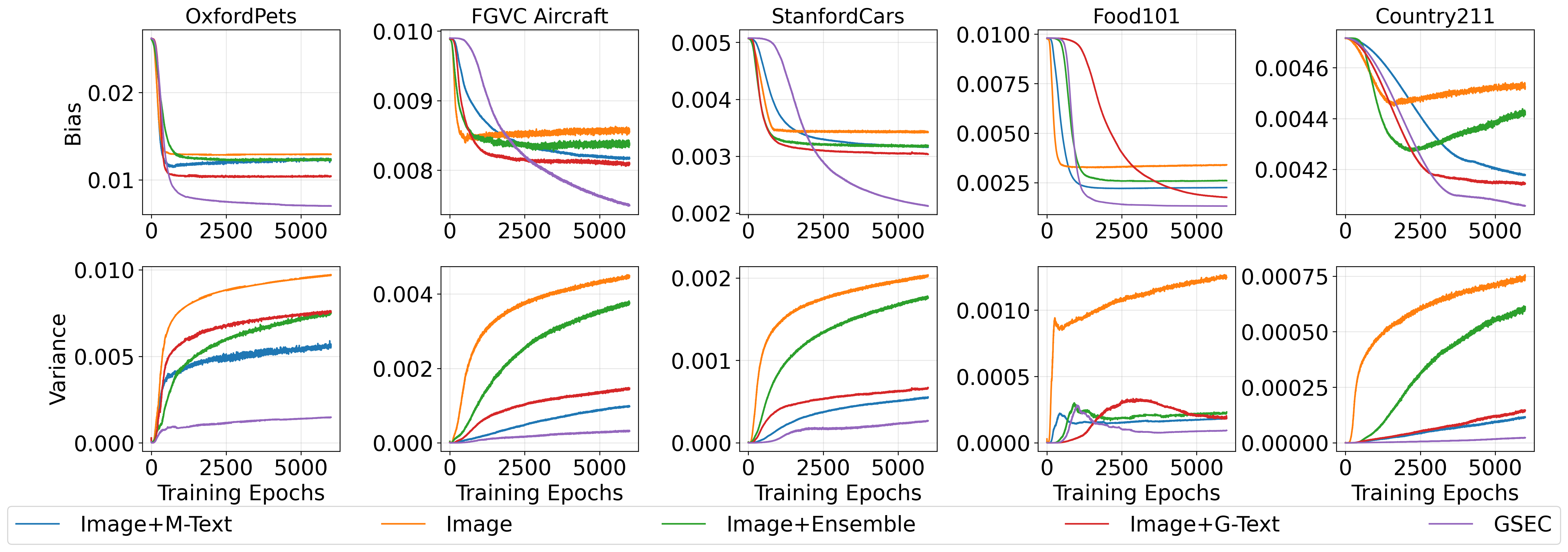}  
	\caption{\textbf{Bias–Variance Analysis.} The figure visualizes the evolution of bias (top row) and variance (bottom row) across five datasets. The curves illustrate that GSEC (purple) consistently achieves the lowest bias and variance compared to other variants.}
	\label{fig:Bias}
\end{figure*}

To evaluate the effectiveness of the proposed method in reducing bias and variance, we first generate 10 base datasets via resampling with the same sample size for each dataset. Each of these datasets is used to train a model. Bias is defined as the average deviation between the model's predictions and the ground truth, while variance quantifies the dispersion across predictions from individual models.

We test the bias-variance performance of the following five configurations. \textbf{Image}: using only image features with the bi‑layer linear architecture; \textbf{Image+Ensemble}: using only image features with the bi‑layer ensemble framework; \textbf{Image+M‑Text}: feeding image features and matching-based textual features into a bi‑layer linear architecture; \textbf{Image+G‑Text}: feeding image features and generative textual features into the bi‑layer linear architecture; \textbf{GSEC}: feeding image features and generative textual features into the bi‑layer ensemble framework.

The experimental results are illustrated in Figure \ref{fig:Bias} and Figure \ref{fig:Bias_1} (in Appendix~\ref{app:Bias-Variance Analysis}). The following observations can be observed: (1) Comparing \textbf{Image} and \textbf{Image+Ensemble}, the ensemble strategy effectively reduces variance. (2) Comparing \textbf{Image+M‑Text} and \textbf{Image+G‑Text}, generative textual guidance helps reduce bias but introduces additional variance. (3) Comparing \textbf{GSEC} with the other methods, GSEC achieves simultaneous reduction of both bias and variance through the complementary effects of generative textual guidance and the ensemble strategy, leading to superior and more stable clustering performance.

\subsection{Ablation Analysis}


To verify the effectiveness of each component in the proposed method, we conduct ablation studies to examine the impact of generative semantic embeddings and the bi‑layer ensemble design on the performance of GSEC. Experiments are carried out across multiple pre‑trained backbones, including CLIP‑ViT‑B/32, CLIP‑RN50x4, CLIP‑RN101, and CLIP‑RN50. Performance is evaluated using Accuracy.

\paragraph{Effectiveness of the Generative Semantic}


We derive semantic prior information from two distinct sources: the Llama-3.2-Vision-11B  MLLM for generative semantics (denoted as G‑Text) and the WordNet for matching semantics (denoted as M‑Text). These textual descriptions are encoded by separate feature extractors and subsequently processed through the same bi‑layer ensemble framework for clustering. As shown in Table~\ref{tab:multimodal_ablation}, G-text consistently outperforms M-text across all tested backbones.

\paragraph{Effectiveness of the Bi-layer Ensemble}


We compare the proposed bi‑layer ensemble framework with a bi‑layer linear architecture under identical image feature inputs. As reported in Table~\ref{tab:ensemble_ablation}, our ensemble‑based approach yields consistent performance gains on ten benchmark datasets across the four different backbones.

\subsection{Hyperparameter Analysis}

\begin{figure}[t]
	\centering
	\includegraphics[width=1.0\linewidth,height=0.18\textheight]{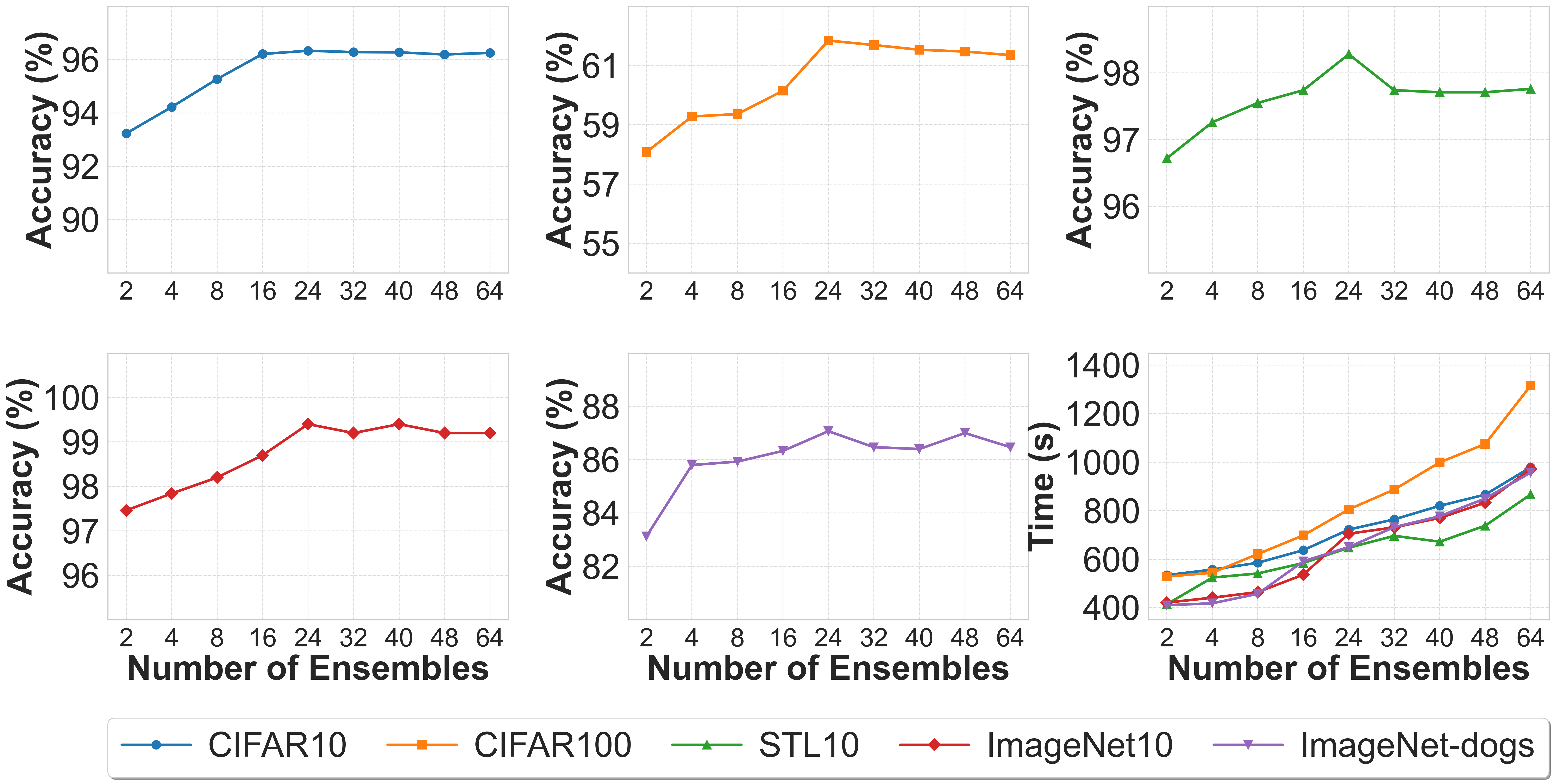}
	\caption{\textbf{Sensitivity analysis of the ensemble size.} The figure evaluates the influence of the number of ensemble members on clustering accuracy (first five subplots) and computational time (last subplot) across five benchmark datasets.}
	\label{fig:ensemble}
\end{figure}

\textbf{Analysis of the number of ensembles.} We analyze the impact of the ensemble size on clustering performance and computational efficiency. As illustrated in Figure~\ref{fig:ensemble}, the runtime increases with the number of ensemble members, and an ensemble size of 24 achieves a favorable balance between performance gains and computational cost.

\begin{figure}[t]
      \centering
       \subfloat[ACC]{\includegraphics[width=0.23\textwidth]{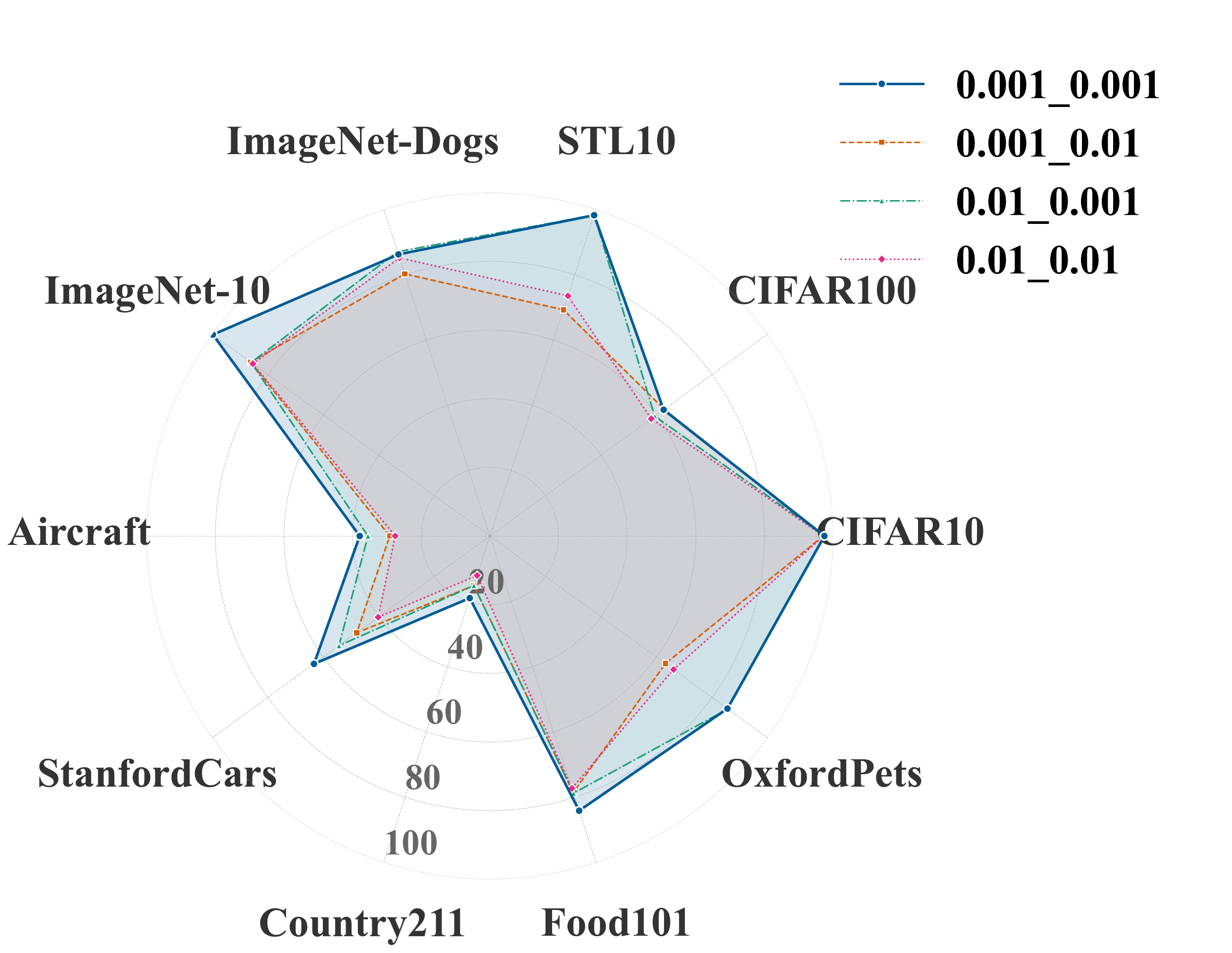}}
       \subfloat[NMI]{\includegraphics[width=0.23\textwidth]{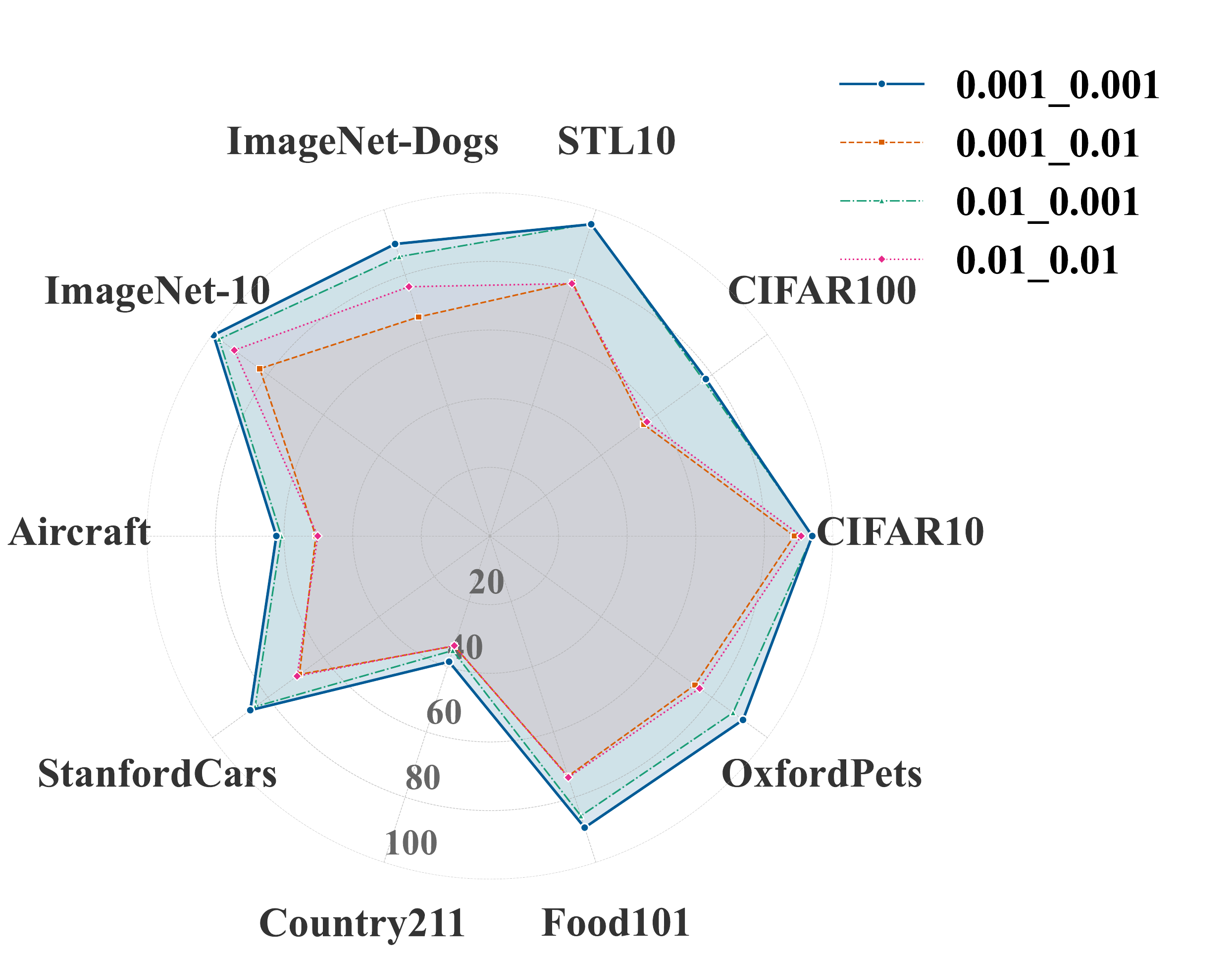}}
\caption{\textbf{Sensitivity analysis of learning rates.} The radar chart illustrates the clustering performance across ten benchmark datasets under four different combinations of inner and outer learning rates.}
\label{fig:Internal and External Learning Rates}
\end{figure}

\textbf{Analysis of Internal and External Learning Rates.}
We investigate the effect of different combinations of inner and outer learning rates on clustering performance, aiming to identify an optimal configuration that ensures stable clustering accuracy. As shown in Figure~\ref{fig:Internal and External Learning Rates}, the model attains the best overall performance across all ten benchmark datasets when both the inner and outer learning rates are set to $0.001$.

\subsection{Convergence Analysis}

\begin{figure}[t]
	\centering
	\includegraphics[width=1.0\linewidth,height=0.08\textheight]{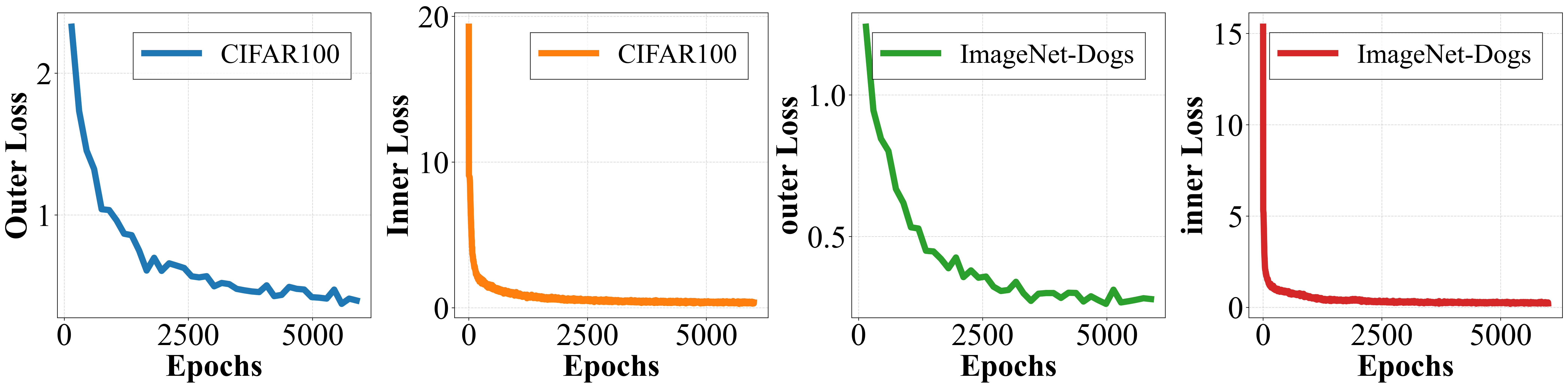}
\caption{\textbf{Convergence analysis of GSEC.} The figure shows the training curves of the outer and inner losses on CIFAR‑100 and ImageNet‑Dogs datasets, illustrating stable optimization behavior.}
	\label{fig:loss}
\end{figure}

To examine the training convergence of GSEC, we visualize the evolution of both inner and outer losses over training epochs. As shown in Figure~\ref{fig:loss}, both losses exhibit a clear downward trend and eventually converge, demonstrating stable optimization.


\subsection{Practicality Analysis}

To assess the efficiency of semantic querying, we report the total query tokens and average latency on CIFAR10 and STL10. As shown in Table~\ref{tab:text_results}, our method incurs limited query overhead and achieves practical response efficiency, with latency of only a few seconds.

\vspace{-0.30cm}
\begin{table}[htbp]
	\centering
	\caption{\textbf{Evaluation of Query Cost and Average Latency.}}
	\begin{minipage}{\columnwidth}
		\resizebox{\columnwidth}{!}{%
			\large         
			\renewcommand{\arraystretch}{1.1}
			\begin{tabular}{l c c|l c c}
				\toprule
				\multirow{2}{*}{CIFAR10} & Query (Tokens) &  Latency (S) & \multirow{2}{*}{STL10} & Query (Tokens) & Latency (S) \\
				& 428986 & 4.07 &  & 48036 & 4.80 \\
				\bottomrule
			\end{tabular}%
		}%
	\end{minipage}
	\label{tab:text_results}
\end{table}
\vspace{-0.33cm}

\section{Conclusion}

This paper proposes GSEC (Image Clustering based on Generative Semantic Guidance and Bi‑Layer Ensemble), a novel framework that jointly reduces bias and variance in image clustering. To overcome the limitations of vocabulary‑matching‑based semantics, GSEC leverages Multimodal Large Language Models to generate adaptive semantic descriptions, which mitigate semantic inconsistency and reduce bias. Simultaneously, a bi‑layer ensemble mechanism is introduced to lower variance and improve robustness. This ensemble consists of an inner layer that integrates cross‑modal information via BatchEnsemble and an outer layer that aligns inner and outer predictions through an alignment‑based ensemble. Experiments on 6 benchmark datasets demonstrate the effectiveness of GSEC in image clustering. Further bias‑variance decomposition confirms that GSEC reduces both bias and variance concurrently, highlighting the importance of jointly optimizing these two factors for improved clustering performance.

\section*{Acknowledgement}
This work was supported by the National Natural Science Foundation of China (Nos. 62441239, 62476160, T2495251, 62441239, 62306170, U24A20253, 62136005), the Special Fund for Science and Technology Innovation Teams of Shanxi Province (No. 202304051001001).
\bibliographystyle{named}
\bibliography{ijcai26}

\newpage
\appendix
\onecolumn

\section{Additional Reduction of Bias-Variance Analysis}\label{app:Bias-Variance Analysis}

\begin{figure}[h!]
	\centering
	\includegraphics[width=0.95\linewidth]{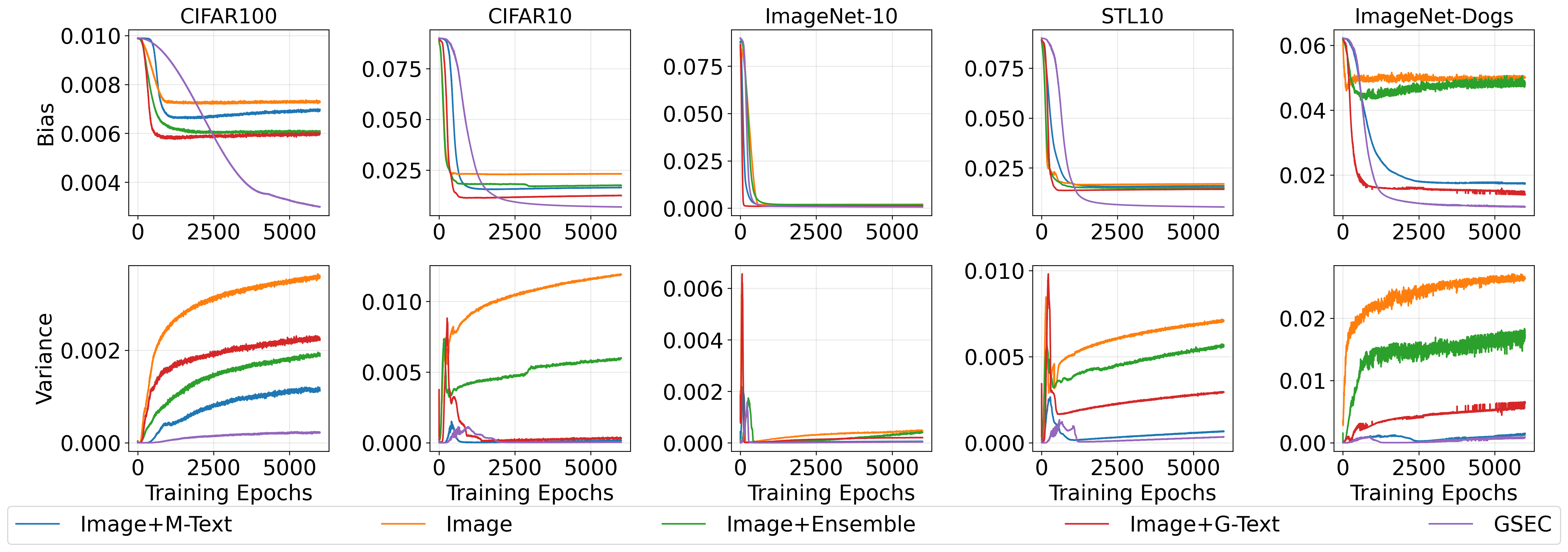}
	\caption{\textbf{Bias–Variance Analysis.} The figure visualizes the evolution of bias (top row) and variance (bottom row) across five datasets. The curves illustrate that GSEC (purple) consistently achieves the lowest bias and variance compared to other variants.}
	\label{fig:Bias_1}
\end{figure}

This section presents bias-variance results on the other five datasets, as illustrated in Figure~\ref{fig:Bias_1}. As illustrated in Figure~\ref{fig:Bias_1}, the results show that: (1) compared with Image, Image+Ensemble exhibits a decrease in variance, indicating that the ensemble strategy effectively reduces variance; (2) compared with Image+M-Text, Image+G-Text significantly reduces bias but increases variance, suggesting that generative textual guidance lowers bias while potentially introducing additional variance; (3) GSEC, which combines generative textual guidance with the ensemble strategy, simultaneously reduces both bias and variance across all comparisons. These observed patterns are consistent with the findings reported in the main text, further demonstrating the effectiveness and strong generalization capability of our method across multiple datasets.

\section{Visualization Analysis}\label{app:Visualized results}

\begin{figure}[!htbp]
	\centering
	\begin{subfigure}[b]{0.19\textwidth}
		\centering
		\includegraphics[width=\linewidth, height=2.5cm, keepaspectratio]{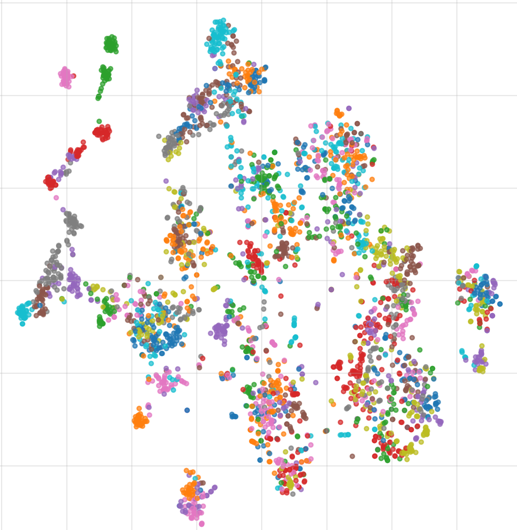}
		\caption{Aircraft}
	\end{subfigure}
	\hfill
	\begin{subfigure}[b]{0.19\textwidth}
		\centering
		\includegraphics[width=\linewidth, height=2.5cm, keepaspectratio]{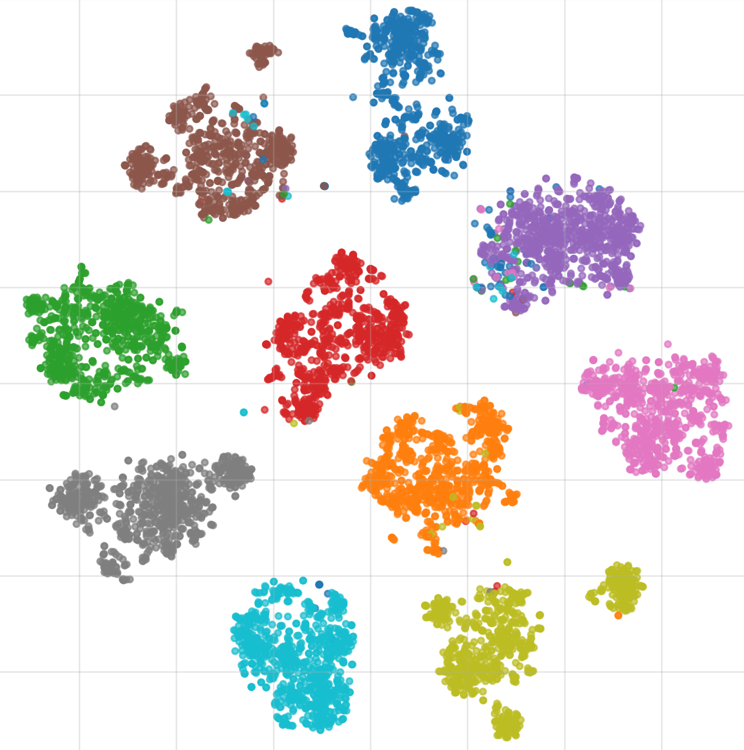}
		\caption{STL10}
	\end{subfigure}
	\hfill
	\begin{subfigure}[b]{0.19\textwidth}
		\centering
		\includegraphics[width=\linewidth, height=2.5cm, keepaspectratio]{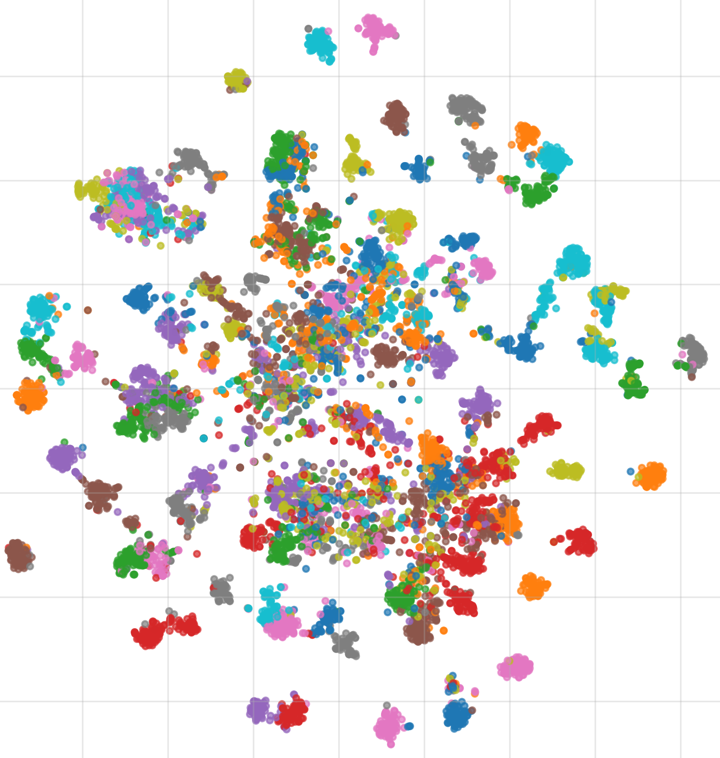}
		\caption{CIFAR100}
	\end{subfigure}
	\hfill
	\begin{subfigure}[b]{0.19\textwidth}
		\centering
		\includegraphics[width=\linewidth, height=2.5cm, keepaspectratio]{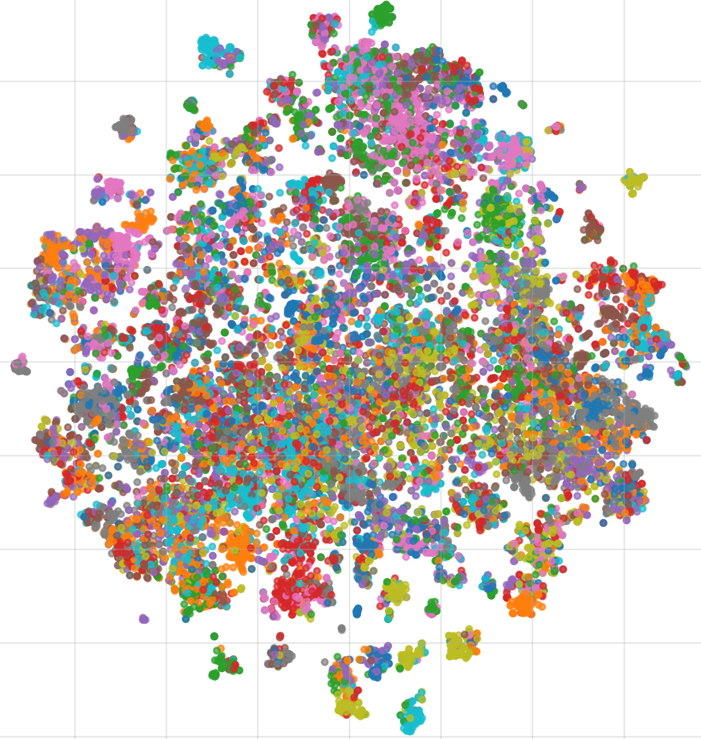}
		\caption{Country211}
	\end{subfigure}
	\hfill
	\begin{subfigure}[b]{0.19\textwidth}
		\centering
		\includegraphics[width=\linewidth, height=2.5cm, keepaspectratio]{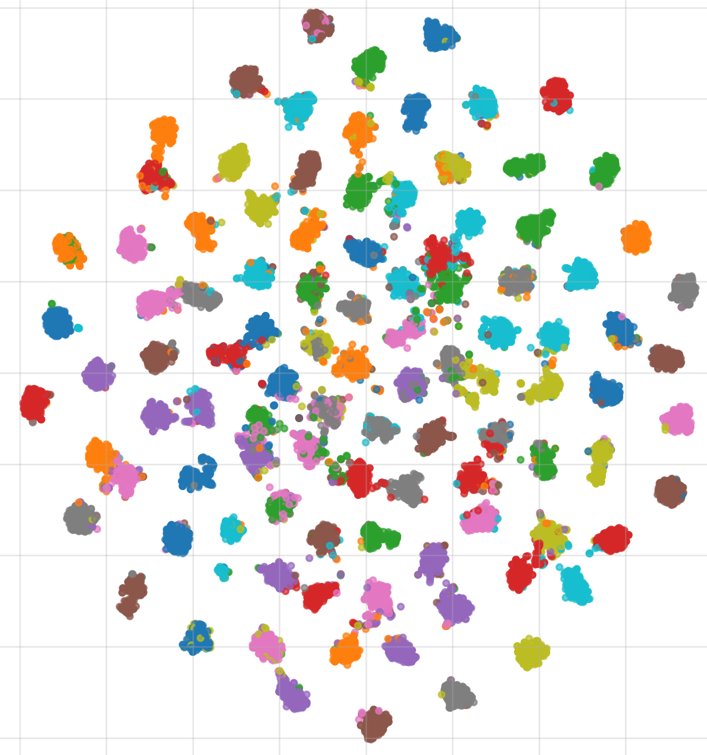}
		\caption{Food101}
	\end{subfigure}
	\begin{subfigure}[b]{0.19\textwidth}
		\centering
		\includegraphics[width=\linewidth, height=2.5cm, keepaspectratio]{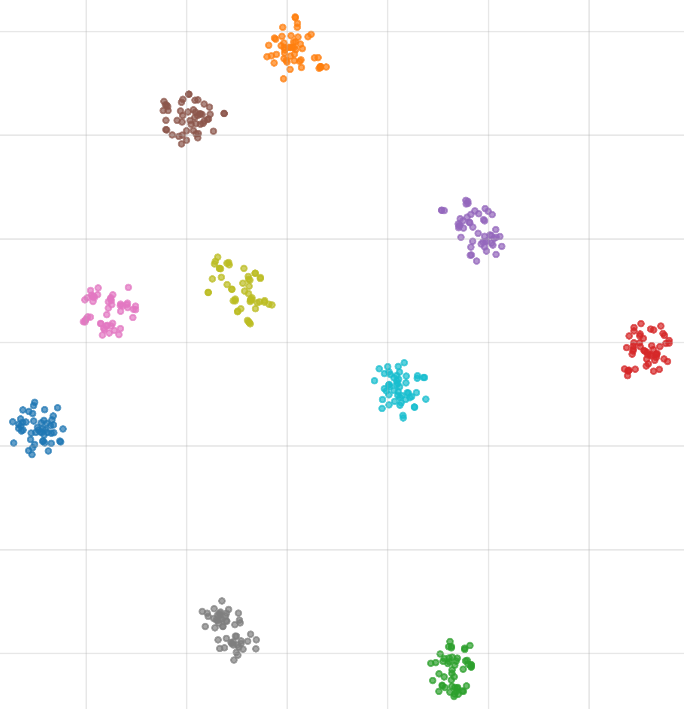}
		\caption{ImageNet-10}
	\end{subfigure}
	\hfill
	\begin{subfigure}[b]{0.19\textwidth}
		\centering
		\includegraphics[width=\linewidth, height=2.5cm, keepaspectratio]{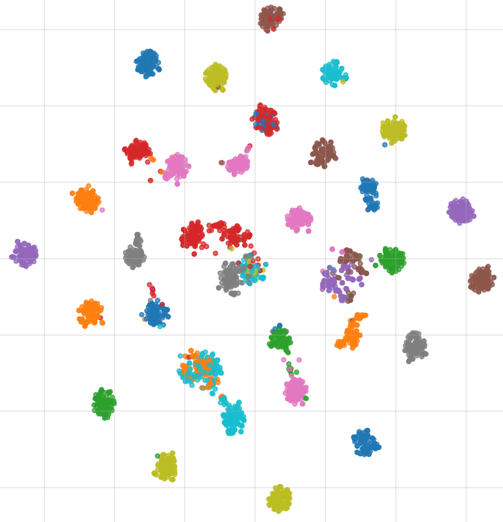}
		\caption{Pets}
	\end{subfigure}
	\hfill
	\begin{subfigure}[b]{0.19\textwidth}
		\centering
		\includegraphics[width=\linewidth, height=2.5cm, keepaspectratio]{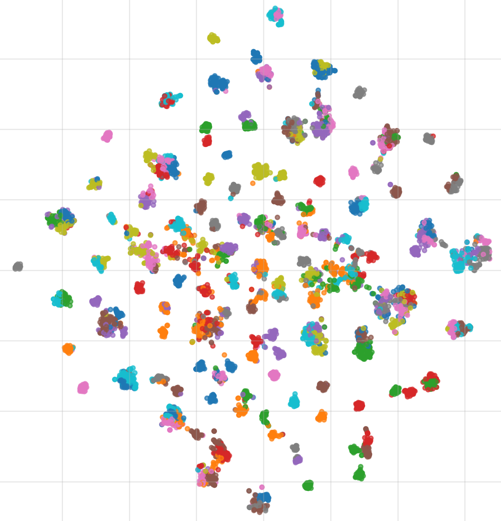}
		\caption{StanfordCars}
	\end{subfigure}
	\hfill
	\begin{subfigure}[b]{0.19\textwidth}
		\centering
		\includegraphics[width=\linewidth, height=2.5cm, keepaspectratio]{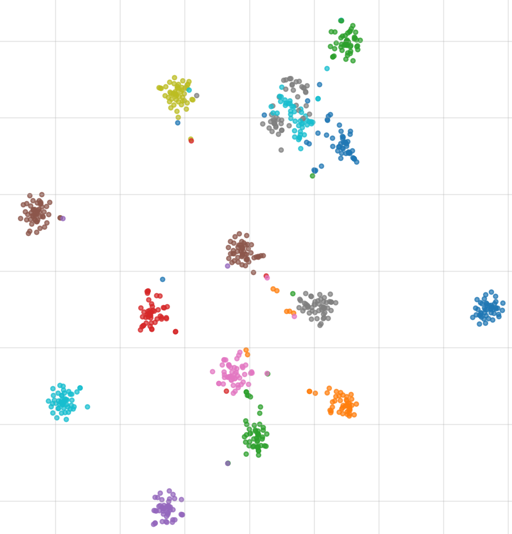}
		\caption{ImageNet-Dogs}
	\end{subfigure}
	\hfill
	\begin{subfigure}[b]{0.19\textwidth}
		\centering
		\includegraphics[width=\linewidth, height=2.5cm, keepaspectratio]{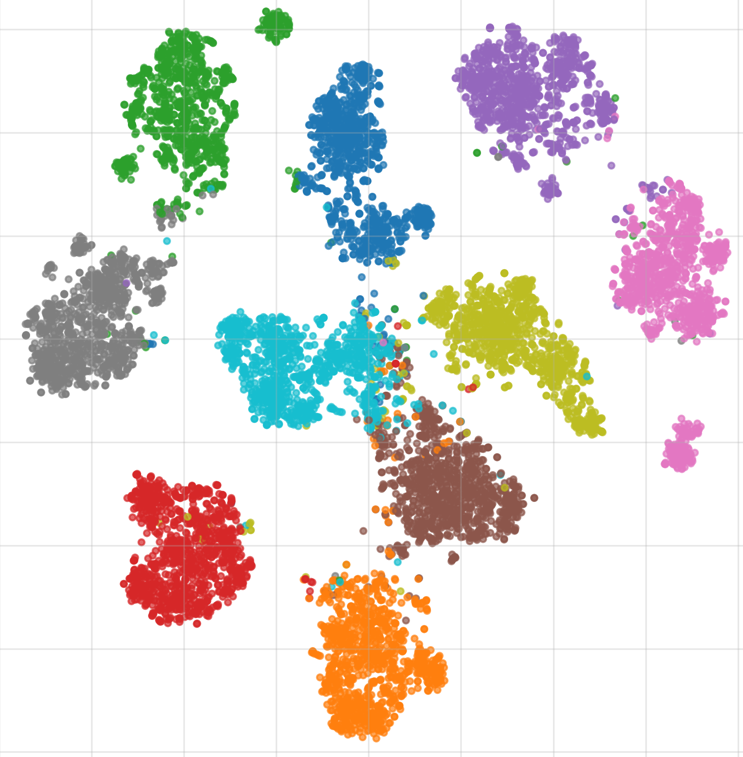}
		\caption{CIFAR10}
	\end{subfigure}
	\caption{\textbf{t-SNE visualization of feature representations.} t-SNE plots illustrate the learned embeddings of GSEC on ten benchmark datasets, showing well-separated clusters and confirming the method's effectiveness.}
	\label{fig:tsne_visualization}
\end{figure}

To visually evaluate the performance of GSEC, we present t-SNE visualizations of the clustering results on ten datasets. As shown in Figure~\ref{fig:tsne_visualization}, the learned features exhibit high intra-class compactness and low inter-class coupling, indicating the effectiveness of the bi-layer ensemble in structuring the feature space. 

\begin{figure}[t!]
	\centering
	\begin{subfigure}[b]{0.235\textwidth}
		\centering
		\includegraphics[width=\linewidth, height=3.6cm]{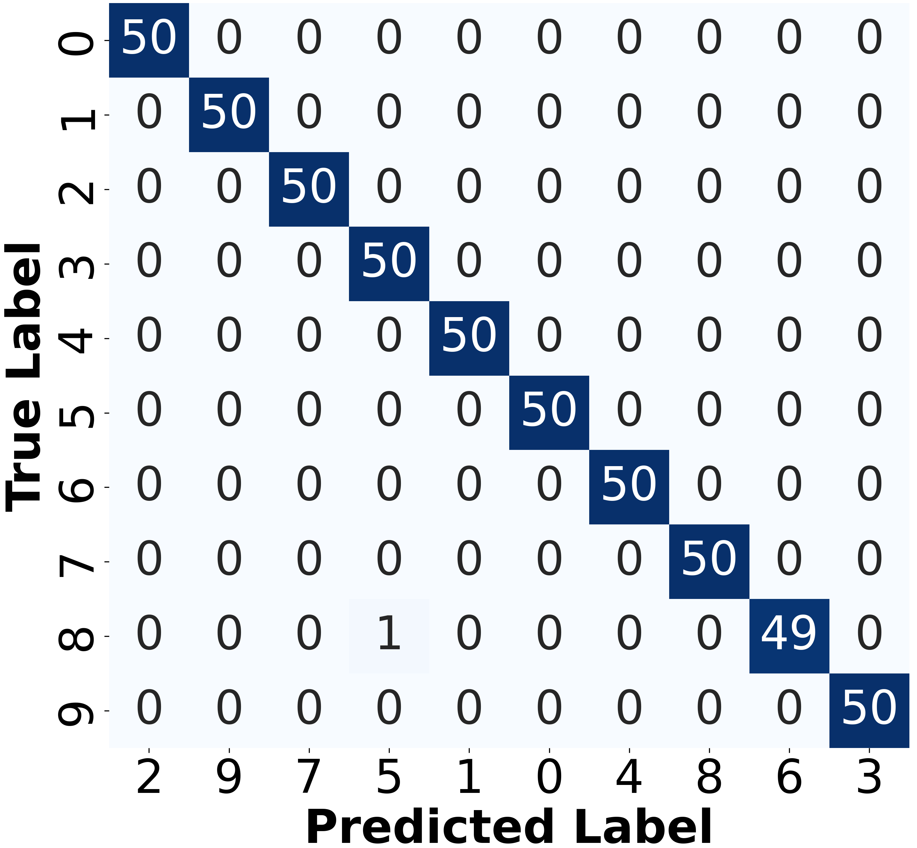}
		\caption{ImageNet-10}
	\end{subfigure}
	\hfill
	\begin{subfigure}[b]{0.235\textwidth}
		\centering
		\includegraphics[width=\linewidth, height=3.6cm]{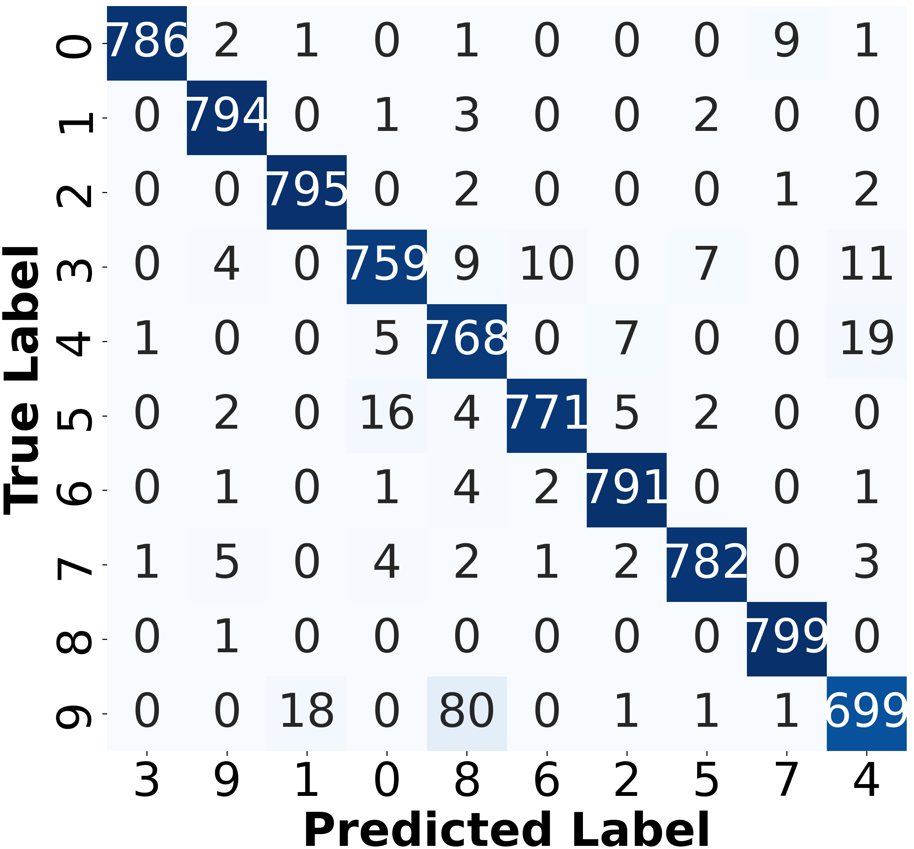}
		\caption{STL10}
	\end{subfigure}
	\hfill
	\begin{subfigure}[b]{0.235\textwidth}
		\centering
		\includegraphics[width=\linewidth, height=3.6cm]{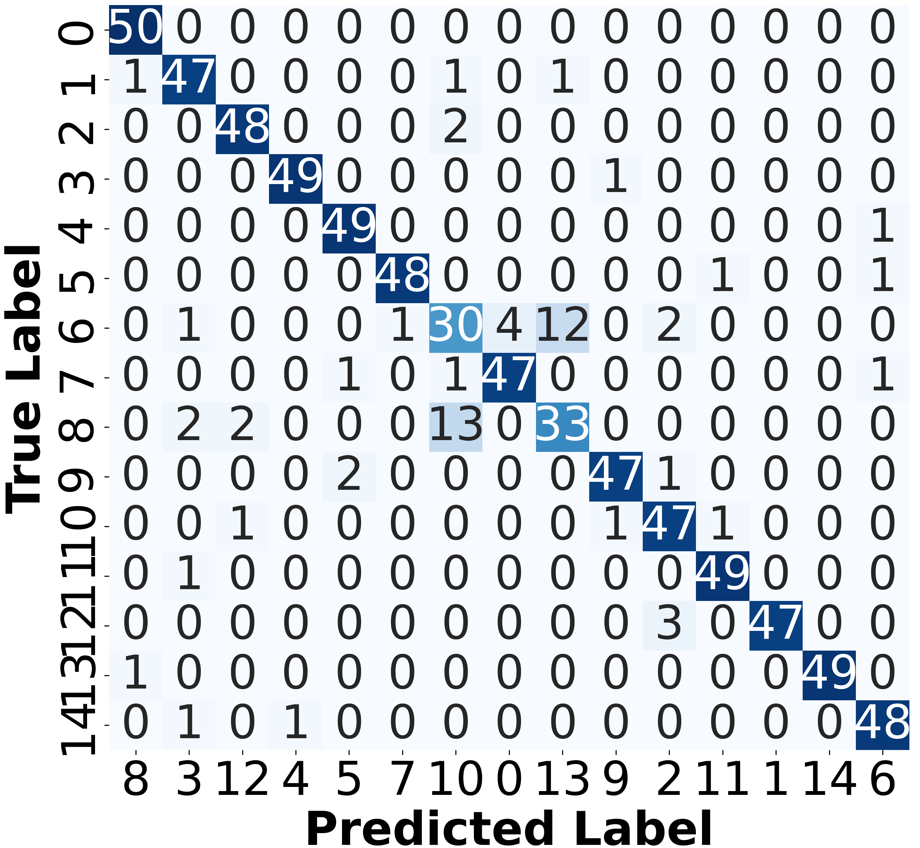}
		\caption{ImageNet-Dogs}
	\end{subfigure}
	\hfill
	\begin{subfigure}[b]{0.235\textwidth}
		\centering
		\includegraphics[width=\linewidth, height=3.6cm]{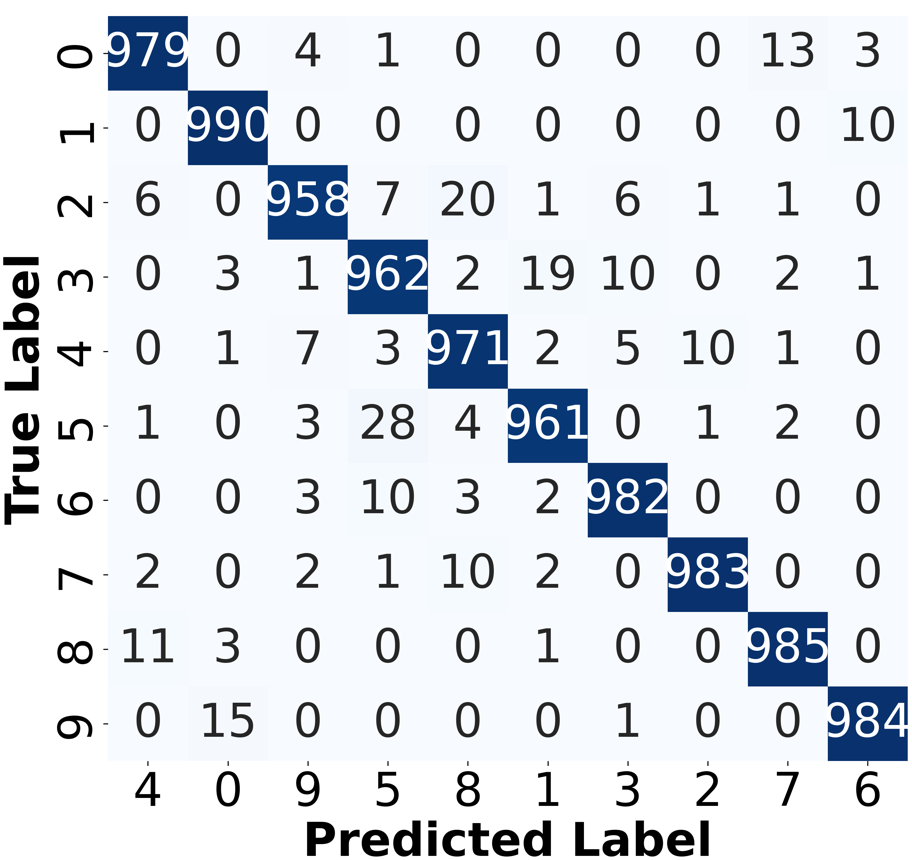}
		\caption{CIFAR10}
	\end{subfigure}
	\caption{\textbf{Confusion matrices.} The figure presents the prediction results of GSEC on ImageNet‑10, STL‑10, ImageNet‑Dogs, and CIFAR‑10. The pronounced diagonal dominance indicates high prediction accuracy across diverse datasets.}
	\label{fig:confusion_matrices}
\end{figure}

Additionally, for datasets with fewer than 30 classes, we provide confusion matrices. The clear diagonal structure shown in Figure~\ref{fig:confusion_matrices} further confirms the accuracy and reliability of the cluster assignments. Together, these visualizations demonstrate that GSEC not only captures the underlying semantic structure of the data accurately but also produces high-quality clustering results with well-defined boundaries.

\clearpage
\makeatletter
\setlength{\@fptop}{0pt}
\makeatother

\end{document}